\documentclass{article}
\usepackage{spconf,amsmath,graphicx}
\usepackage{tikz}
\usetikzlibrary{shapes.geometric, arrows}
\usepackage{multirow}

\usepackage{color}
\usepackage[accsupp]{axessibility}  %

\usepackage{microtype}
\usepackage{adjustbox}
\usepackage{listings}
\usepackage{pythonhighlight}
\usepackage{float}
\usepackage{tablefootnote}
\usepackage{tabularx}
\usepackage{booktabs}
\usepackage{graphicx}
\usepackage{float}
\usepackage{wrapfig}

\usepackage{graphicx}
\usepackage{amsmath}
\usepackage{amssymb}
\usepackage{booktabs}
\usepackage{textgreek}

\usepackage[capitalize]{cleveref}
\crefname{section}{Sec.}{Secs.}
\Crefname{section}{Section}{Sections}
\Crefname{table}{Table}{Tables}
\crefname{table}{Tab.}{Tabs.}

\newcommand{\MAE}{$\mathsf{CSMAE~}$}
\DeclareSymbolFont{Xlargesymbols}{OMX}{cmex}{m}{n}
\DeclareMathSymbol{\Xsum}{\mathop}{Xlargesymbols}{80}

\title{\MAE:~Cataract Surgical Masked Autoencoder (MAE) based Pre-training}
\name{%
\begin{tabular}{@{}c@{}}
Nisarg A. Shah$^{1}$ \qquad
Wele Gedara Chaminda Bandara$^{1}$ \qquad
Shameema Skider$^{2,3}$\\
S. Swaroop Vedula$^{3}$ \qquad
Vishal M. Patel$^{1}$
\end{tabular}
}                                                                                
\address{
$^{1}$ Johns Hopkins University, Baltimore, USA \\
$^{2}$ Wilmer Eye Institute, Johns Hopkins University, Baltimore, USA \\
$^{3}$ Malone Center for Engineering in Healthcare, Johns Hopkins University, Baltimore, USA
}

\begin{document}
\maketitle
\begin{abstract}
Automated analysis of surgical videos is crucial for improving surgical training, workflow optimization, and postoperative assessment. 
We introduce a \MAE, Masked Autoencoder (MAE)-based pretraining approach, specifically developed for Cataract Surgery video analysis, where instead of randomly selecting tokens for Masking, they are selected based on spatiotemporal importance of the token. 
We created a large dataset of cataract surgery videos to improve the model's learning efficiency and expand its robustness in low-data regime.
Our pre-trained model can be easily adapted for specific downstream tasks via fine-tuning, serving as a robust backbone for further analysis. 
Through rigorous testing on downstream step—recognition task on two Cataract surgery video datasets, D99 and Cataract-101—our approach surpasses current state-of-the-art self-supervised pre-training and adapter-based transfer learning methods by a significant margin. This advancement not only demonstrates the potential of our MAE-based pretraining in the field of surgical video analysis but also sets a new benchmark for future research.
\end{abstract}
\section{Introduction}

Recent advances in medical imaging show the effectiveness of Transformer pre-training on large datasets for classification, detection, and segmentation. Due to the variety of imaging types and high data costs, models tailored to specific data are often necessary \cite{kang2023deblurring,lu2023multi}. Our research focuses on cataract surgical videos, essential for eye disease diagnosis, training, and minimally invasive surgeries.

Medical imaging often combines images and text, as seen in X-ray diagnosis~\cite{boecking2022making} and radiology report generation~\cite{moon2022multi} with vision-language models. However, cataract videos lack text, requiring image/video-only models. Tasks like step recognition need a lot of expert-labeled data, which is hard to obtain\cite{shah2023glsformer}. We propose \MAE, a self-supervised learning approach using masked autoencoders (MAE) to pre-train on unlabeled cataract videos, enabling feature learning for step recognition.

Self-supervised learning includes two main strategies: 1) Contrastive Learning~\cite{hu2021contrast,park2022probabilistic} and 2) Masked Autoencoders (MAEs)~\cite{he2022masked,tong2022videomae}. Contrastive learning compares augmented image versions, clustering similar images and distancing different ones in latent space. MAEs, in contrast, divide an image or video into patches, mask some randomly, and reconstruct them using features from unmasked patches. MAEs use a Vision Transformer (ViT)~\cite{dosovitskiy2020image,arnab2021vivit} encoder for visible patches and a lightweight decoder for prediction.
MAEs are particularly effective, as they learn from incomplete data and outperform contrastive learning on multiple tasks.

We developed a video transformer model to encode patches, inspired by the ViT-B architecture in VideoMAE~\cite{tong2022videomae}. In training, we tested various masking techniques—random, tube, and frame—with random masking showing improved results. However, since not all patches hold equal information, random sampling can overlook key spatiotemporal cues. To address this, our approach selects patches based on their spatiotemporal significance rather than standard random sampling as propsed in \cite{bandara2023adamae} for general vision videos. An auxiliary network defines a categorical distribution across input tokens, from which visible tokens are sampled, ensuring a more focused, informative selection.

In this work, we show the effectiveness of the MAE-based approach for Vision Transformer pretraining in cataract videos with the following contributions:
(1) An end-to-end token sampling strategy that selects high-informative spatiotemporal tokens, discarding redundant ones for efficient MAE pre-training on long cataract surgery videos;  
(2) Improved GPU memory efficiency and enhanced feature learning as the \MAE model focuses on spatiotemporal-rich regions; and  
(3) Extensive evaluation of the \MAE pre-trained model on Step Recognition, demonstrating its effectiveness in both semi-supervised and supervised settings.

\begin{figure*}[htp!]
\centering
  \includegraphics[page=1,width=.5\linewidth]{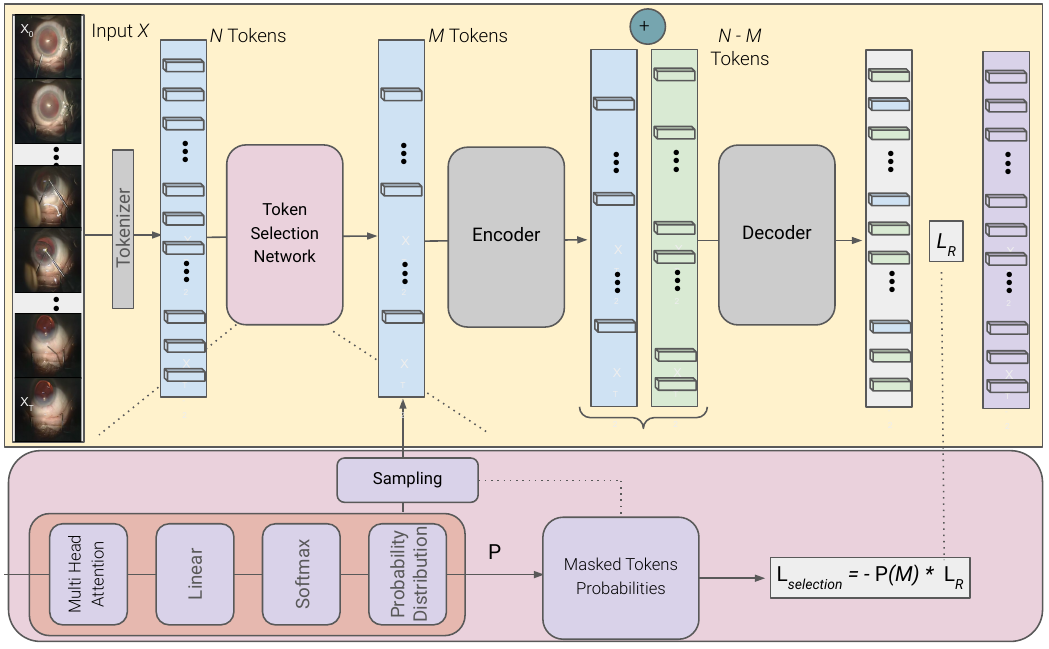}
 \vskip-10pt \caption{{The \MAE model, developed on an encoder-decoder framework, consists of four principal components: a Tokenizer, Token Selection Network, an Encoder, and a Lightweight Decoder. The Tokenizer converts raw video data into a token-based feature representation. Utilizing transformer architecture, the Token Selection Network predicts the probability distribution for each token, choosing those with significant spatiotemporal information to feed into the Encoder. This Encoder, which employs the ViT network, captures the feature representation of these selected tokens. These representations are then combined with learnable feature representations of masked tokens, aiming to reconstruct the masked feature representation accurately.}}
  \label{fig:model_arc}
\end{figure*}
\section{The \MAE~model}

Fig.~\ref{fig:model_arc} shows the \MAE framework similar to~\cite{bandara2023adamae,tong2022videomae}, composed of four main components: a Tokenizer, Encoder, Decoder, and a Token Selection Network.

\textbf{Tokenizer:} For a video input \(X\) of dimensions \(T \times C \times H \times W\) (temporal frames, RGB channels, and spatial dimensions), the Tokenizer~\cite{tong2022videomae} uses a 3D convolution layer to convert \(X\) into \(N\) tokens, each with dimension \(k\), and applies positional encoding~\cite{he2022masked,tong2022videomae}. 

\textbf{Token Selection Network:} Following~\cite{bandara2023adamae}, the Tokenizer output \(T\) is processed through a Multi-Head Attention (MHA) layer, a Linear layer, and Softmax to generate probability scores \(P\) for each token:
\[
P = \text{Softmax}(\text{Linear}(\text{MHA}(T))).
\]
An N-dimensional categorical distribution is applied to \(P\), selecting a subset \(M_i\) of visible tokens based on a masking ratio \(\alpha \in (0, 1)\).

\textbf{Encoder:} The encoder processes selected visible tokens \(T_v\) via a ViT encoder to produce latent representations \(F_v\).

\textbf{Decoder:} The decoder combines \(F_v\) with a fixed representation for masked tokens using positional encoding. A transformer then reconstructs the masked video frames \(X_{n-m}\).

\subsection{Training \MAE}

\textbf{Masked Reconstruction Loss} \(L_R\): We use Mean Squared Error (MSE) loss between predicted masked tokens \(X_b\) and ground-truth normalized RGB values \(X_e\):
\[
L_R(\phi) = \frac{1}{N - M} \sum_{i \in M_i^{'}} \|X_{b_i} - X_{e_i}\|^2,
\]
where \(X_b\) are the predicted tokens, and \(X_e\) is the ground-truth.

\begin{table*}[ht]
\centering
\caption{Comparison of \MAE with other state-of-the-art methods under different data-regime settings on the D99 dataset\cite{shah2023glsformer}.} 
\resizebox{0.5\textwidth}{!}{%
\begin{tabular}{c|c|c|cccc}
\toprule
Methods & Pre-training Dataset & Masking & \multicolumn{4}{c}{Data Regime (\%)} \\
\cmidrule{4-7}
 &  &  & 10 & 25 & 50 & 100 \\
\midrule
MaskFeat~\cite{wei2022masked} & Kinetics-400 & Random & 47.28 & 59.32 & 60.47 & 72.85 \\
GLSFormer~\cite{shah2022towards} & Kinetics-400 & - & 47.19 & 61.54 & 63.76 & \textbf{80.24} \\
VideoMAE~\cite{tong2022videomae} & D-450 & Frame & 48.62 & 58.73 & 60.84 & 70.91 \\
STMAE~\cite{feichtenhofer2022masked} & D-450 & Random & 52.37 & 60.42 & 63.58 & 74.16 \\
VideoMAE~\cite{tong2022videomae} & Kinetics-400 & Tube & 46.16 & 59.76 & 60.99 & 73.35 \\
VideoMAE~\cite{tong2022videomae} & D-450 & Random & 50.24 & 60.89 & 62.34 & 72.98 \\
VideoMAE~\cite{tong2022videomae} & D-450 & Tube & 52.11 & 61.59 & 63.72 & 74.39 \\
\MAE & D-450 & Token Selection & \textbf{54.75} & \textbf{63.12} & \textbf{65.83} & 78.14 \\
\bottomrule
\end{tabular}%
}
\label{tbl:table_1}
\end{table*}

\begin{table*}[h]
\centering
\footnotesize
\caption{Quantitative results of step recognition from different methods on the D99 and Cataract-101 datasets. 
}
\resizebox{0.85\textwidth}{!}{%
\begin{tabular}{ccccccccc} \toprule
\multirow{2}{*}{Method} & \multicolumn{4}{c}{D99}               & \multicolumn{4}{c}{Cataract-101}       \\   \cline{2-5}\cline{6-9} 
                        & Jaccard & Precision & Recall & Accuracy & Jaccard & Precision & Recall & Accuracy  \\ \toprule
ResNet\cite{he2016deep}     & 37.98 \(\pm\) 2.97 & 54.76 \(\pm\) 2.77 & 52.28 \(\pm\) 2.89 & 72.06 \(\pm\) 2.12 & 62.58 \(\pm\) 1.92 & 76.68 \(\pm\) 1.86 & 74.73 \(\pm\) 1.27 & 82.64 \(\pm\) 1.54        \\
SV-RCNet\cite{jin2017sv}    &  39.15 \(\pm\) 2.03 & 58.18 \(\pm\) 1.67 & 54.25 \(\pm\) 1.86 & 73.39 \(\pm\) 1.64 & 66.51 \(\pm\) 1.30 & 84.96 \(\pm\) 0.94 & 76.61 \(\pm\) 1.18 & 86.13 \(\pm\) 0.91     \\
OHFM\cite{yi2019hard}       &  40.01 \(\pm\) 1.68 & 59.12 \(\pm\) 1.33 & 55.49 \(\pm\) 1.63 & 73.82 \(\pm\) 1.13 & 69.01 \(\pm\) 0.93 & 85.37 \(\pm\) 0.78 & 78.29 \(\pm\) 0.81 & 87.82 \(\pm\) 0.71          \\
TeCNO\cite{czempiel2020tecno}       &  41.31 \(\pm\) 1.72 & 61.56 \(\pm\) 1.41 & 55.81 \(\pm\) 1.58 & 74.07 \(\pm\) 1.78 & 70.18 \(\pm\) 1.15 & 86.03 \(\pm\) 0.83 & 79.52 \(\pm\) 0.90 & 88.26 \(\pm\) 0.92        \\
TMRNet\cite{jin2021temporal} &  41.42 \(\pm\) 1.76 & 61.37 \(\pm\) 1.46 & 56.02 \(\pm\) 1.65 & 75.11 \(\pm\) 0.91 & 71.83 \(\pm\) 0.91 & 85.09 \(\pm\) 0.72 & 82.44 \(\pm\) 0.75 & 89.68 \(\pm\) 0.76      \\ 
Trans-SVNet\cite{gao2021trans} &  42.06 \(\pm\) 1.51 & 60.12 \(\pm\) 1.55 & 56.36 \(\pm\) 1.24 & 74.89 \(\pm\) 1.37 & 72.32 \(\pm\) 1.04 & 86.72 \(\pm\) 0.85 & 81.12 \(\pm\) 0.93 & 89.45 \(\pm\) 0.88       \\ %
ViT\cite{dosovitskiy2020image}         &  38.18 \(\pm\) 2.79 & 55.15 \(\pm\) 2.42 & 53.60 \(\pm\) 2.63 & 72.45 \(\pm\) 1.91 & 64.77 \(\pm\) 1.97 & 78.51 \(\pm\) 1.42 & 75.62 \(\pm\) 1.83 & 84.56 \(\pm\) 1.72       \\
TimesFormer\cite{bertasius2021space} &   42.69 \(\pm\) 1.34 & 64.24 \(\pm\) 1.20 & 55.17 \(\pm\) 1.26 & 77.83 \(\pm\) 0.96 & 75.97 \(\pm\) 1.26 & \textbf{85.38 \(\pm\) 0.93} & 84.47 \(\pm\) 0.95 & 90.76 \(\pm\) 1.05       \\
STMAE\cite{feichtenhofer2022masked} &  41.67 $\pm$ 1.29 & 59.38 $\pm$ 2.46 & 53.22 $\pm$ 1.78 & 74.16 $\pm$ 1.39 & 70.54 $\pm$ 1.63 & 81.47 $\pm$ 2.35 & 78.67 $\pm$ 1.54 & 85.29 $\pm$ 1.62       \\
VideoMAE\cite{tong2022videomae} &   42.58 $\pm$ 1.73 & 61.24 $\pm$ 1.20 & 56.35 $\pm$ 1.82 & 74.39 $\pm$ 1.47 & 71.39 $\pm$ 1.25 & 82.13 $\pm$ 1.39 & 80.16 $\pm$ 1.68 & 86.47 $\pm$ 1.52       \\
\MAE   & \textbf{43.51 $\pm$ 1.47} & \textbf{64.32 $\pm$ 1.36} & \textit{52.45 $\pm$ 1.17} & \textbf{78.14 $\pm$ 1.25} & \textbf{76.82 \(\pm\) 2.27} & {84.26 \(\pm\) 1.64} & \textbf{86.73 \(\pm\) 1.38} & {89.83 \(\pm\) 1.15}   \\ \bottomrule
\end{tabular}}
\label{tbl:table_2}
\end{table*}

\textbf{Token Selection Loss} \(L_{select}\): We train the token selection mechanism, parameterized by \(\theta\), using a sampling loss \(L_{select}\) based on the Gradient-Following algorithm~\cite{williams1992simple,bandara2023adamae} from reinforcement learning. This approach treats token selection as actions within an MAE environment, with the masked reconstruction loss \(L_R\) as the return. The goal is to maximize the expected reconstruction error \(E[L_R]\), akin to maximizing expected rewards.

Following compressed sensing principles, tokens are sampled more heavily in high-activity regions, allowing for masking ratios up to 95\% and leading to efficient sampling and reduced computational load. This strategy enables faster pre-training by concentrating tokens in high-information areas.

The objective function for optimizing token selection is:
\[
L_{select}(\theta) = -E_\theta[L_R(\phi)] = -\sum_{i \in I_m} P_{i\theta} \cdot L_{iR}(\phi),
\]
where \(P_{i\theta}\) is the token selection probability for index \(i\), and \(L_{iR}\) is the reconstruction error for each masked token~\cite{bandara2023adamae}. To isolate MAE parameters (\(\phi\)) from \(L_{select}(\theta)\), gradient updates are detached, and logarithmic scaling is applied to probabilities to address precision issues.

\section{Experiments and Results}

\begin{table*}[ht]
\centering
\caption{Ablation studies on step-recognitions in semi-supervised setting (10\% labeled data) using Video-Vision Transformer as a backbone. (a) Different Masking Ratio: \MAE works well with extremely high masking ratio, models are trained for 800 epochs. (b) Decoder Depth: \MAE performs the best with 4 blocks of decoder, models are trained for 800 epochs with a masking ratio of 95\%. (c) Mask sampling: \MAE outperforms random, frame, and tube masking. (d) Pre-training epochs: Better performance is achieved during fine-tuning when pre-trained for more epochs. (e) Loss function: \MAE performs best with MSE loss (normalized) and $L_S$, models are trained for 800 epochs with a masking ratio of 95\%.}
\resizebox{0.7\textwidth}{!}{%
\begin{tabular}{cc|cc|cc|cc|cc}
\toprule
 Ratio & mAP & Blocks & mAP & Case & mAP & Epochs & mAP & Case & mAP \\
\midrule
 0.98 & 52.47 & 1 & {50.69} & Random & 50.53 & 200 & 50.28 & L1 (w / norm)  & 53.62 \\
 0.95 & \textbf{54.75} & 2 & 52.34 & Tube & 52.19 & 400 & 52.42 & L1 (wout / norm) & 52.56 \\
 0.90 & 53.85 & 4 & \textbf{54.75} & Frame & 43.92 & 600 & 53.97 & MSE (w / norm) & \textbf{54.75} \\
 0.80 & 50.82 & 8 & 53.68 & \MAE & \textbf{54.75} & 800 & \textbf{54.75} & MSE (wout / norm) & 52.89 \\
\bottomrule
\end{tabular}%
}
\label{tab:ablation_revised}
\end{table*}

\noindent {\bf{Datasets.}} For pre-training, we extend the D99 dataset to create D450, adding 350 cataract surgery videos for a total of 450 videos, each averaging 34 minutes at 59 \textit{fps}. We evaluate \MAE on two untrimmed cataract datasets: Cataract-101~\cite{schoeffmann2018cataract} and D99~\cite{yu2019assessment}.
For Step Recognition fine-tuning, we experiment with both low-data and full-data settings. The D99 dataset has 99 videos with 12 annotated steps, a $640\times480$ resolution, and 59 \textit{fps}. Following~\cite{shah2023glsformer}, we split D99 into 60, 20, and 19 videos for training, validation, and testing. Cataract-101 includes 101 videos at 25 \textit{fps} with 10 annotated steps and a $720\times540$ resolution, divided into 50, 10, and 40 videos for training, validation, and testing~\cite{shah2023glsformer}.
All videos are subsampled to 1 fps and resized to $250\times250$ for pre-training and fine-tuning, as in prior studies~\cite{gao2021trans,twinanda2016endonet}.

\noindent {\bf{Evaluation Metrics.}}
To compare the results of \MAE on step recognition, following \cite{shah2023glsformer,kim2019objective}
we estimated accuracy, precision, recall, and Jaccard index. \\

\begin{figure}[htp!]
\centering
  \includegraphics[page=2,width=0.75\linewidth]{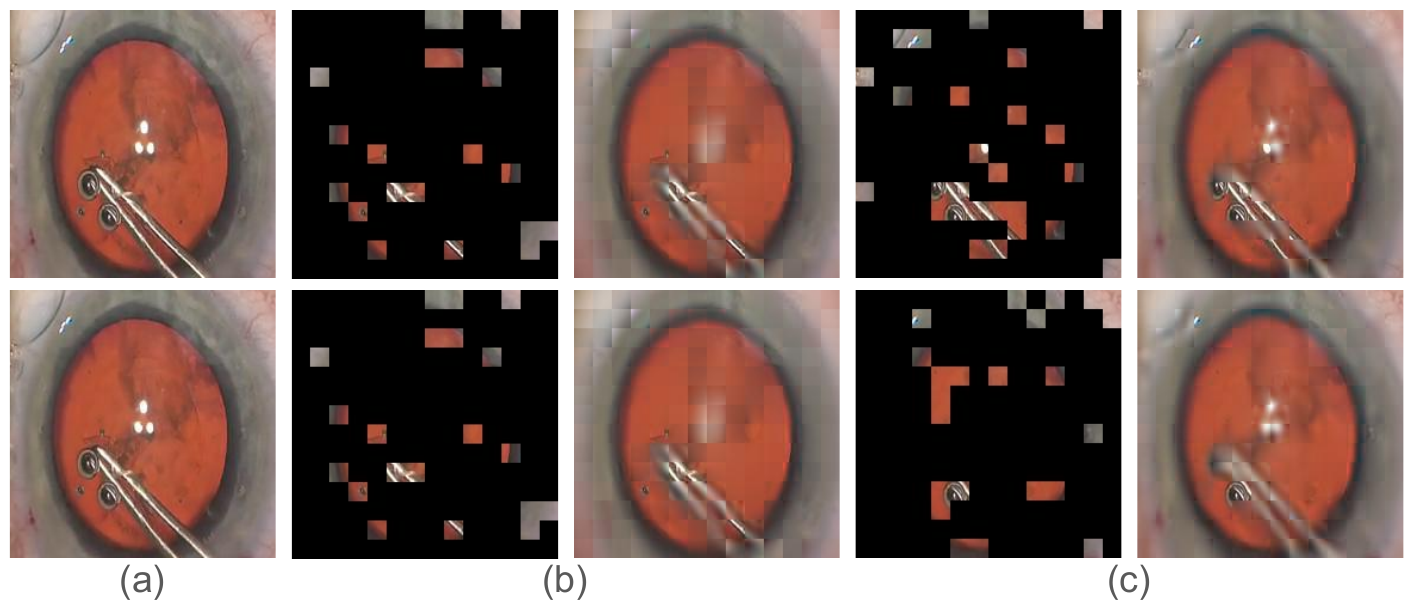}
 \vskip-10pt \caption{Qualitative Comparison of Reconstructed Images after Pre-training Experiment. (a) indicates Original Images, (b) indicates Mask and Reconstructed Image from VideoMAE\cite{tong2022videomae} and (c) indicates Selected Masks from TSN and Reconstructed Image from \MAE.}
  \label{fig:results}
\end{figure}

\noindent \textbf{Comparison with state-of-the-art methods.}
Table~\ref{tbl:table_1} compares our \MAE model with state-of-the-art pre-training and fine-tuning methods across multiple data-regime settings on the D99 dataset. Although VideoMAE~\cite{tong2022videomae} was trained with 90\% masking, our \MAE achieves consistently strong performance with a 95\% masking ratio. Against other pre-training methods like MaskFeat~\cite{wei2022masked}, STMAE~\cite{feichtenhofer2022masked}, and variations of VideoMAE, we observe gains of up to 7\% with fewer data.

In both high- and low-data regimes, our model performs comparably or better than specialized step recognition models like GLSFormer~\cite{shah2023glsformer}. Notably, at 10\% labeled data, \MAE shows an improvement of about 8\%, reflecting the effectiveness of the Token Selection Network in focusing on high-informative spatiotemporal regions. 

Furthermore, our model demonstrates strong transferability on the Cataract-101 dataset (Table~\ref{tbl:table_2}), achieving up to 3–5\% improvements in Jaccard and recall over pre-training methods like STMAE, VideoMAE, and specialized networks such as Trans-SVNet and TimesFormer, highlighting the generalizability of the learned features across different surgical settings.

\noindent \textbf{Qualitative Results.}
Figure~\ref{fig:results} compares VideoMAE~\cite{tong2022videomae} and our \MAE approach, illustrating masked tokens and frame reconstruction quality in cataract videos. VideoMAE (b) uses a consistent tube masking across frames, whereas our \MAE (c) applies a dynamic, frame-specific masking pattern, enhancing the capture of spatial and temporal details.

Despite VideoMAE's lower masking ratio (90\%), its reconstructed frames show noticeable blurriness, while our method, even with a 95\% masking ratio, preserves finer details like cataract textures and surgical bubbles, indicating higher fidelity in critical feature representation.

Additionally, \MAE's mask distribution aligns with non-uniform sampling, prioritizing tokens from high-information regions, such as surgical instruments and cataract morphology, over the static background, demonstrating its ability to focus on regions with significant spatiotemporal changes.

\noindent \textbf{Ablation Studies.}
We ablate the \MAE design by evaluating its performance on the VVT architecture for semi-supervised step recognition with 10\% labeled data. Our analysis focuses on masking strategies, decoder depths, masking ratios, and loss functions.

\textit{Masking Strategies:}
Table~\ref{tab:ablation_revised}c shows that random patch masking (50.52\%) and tube masking~\cite{tong2022videomae} (51.19\%) outperform frame-based masking (43.92\%). Adaptive token sampling, maximizing reconstruction error, achieves the highest mAP (54.75\%) with a 95\% masking ratio, surpassing random masking at 90\%.

\textit{Decoder Depth:}
As shown in Table~\ref{tab:ablation_revised}b, increasing decoder blocks from 1 to 4 improves mAP from 50.69\% to 54.75\%, with performance declining beyond four blocks, suggesting an optimal depth of 4 for balancing complexity and efficiency, consistent with VideoMAE~\cite{tong2022videomae}.

\textit{Loss Function and Training Epochs:}
Table~\ref{tab:ablation_revised}e indicates L1/MSE loss on normalized pixels outperforms raw pixel loss. Table~\ref{tab:ablation_revised}d shows a 4.5\% accuracy improvement by extending pretraining epochs from 200 to 800, highlighting the benefits of longer training.

\section{Conclusion} %
In this paper, we present a novel masking technique, \MAE, tailored for Masked Autoencoder (MAE)-based pretraining on cataract surgical videos to enhance spatiotemporal representation learning. Our approach, designed for Video Vision Transformer models, uses a pretraining dataset of over 350 untrimmed surgical videos (avg. 34 minutes each) to prepare models for fine-tuning on downstream tasks with minimal labels.
Qualitative analysis shows that our sampling method surpasses competing masking strategies in both supervised and semi-supervised settings on cataract surgery tasks, highlighting its potential for broader clinical applications.
\\
\section{Acknowledgments}
This research was supported by a grant from the National Institutes of Health, USA; R01EY033065. The content is solely the responsibility of the authors and does not necessarily represent the official views of the National Institutes of Health.

\bibliographystyle{IEEEbib}
\bibliography{refs}

\end{document}